\documentclass[conference]{IEEEtran}
\IEEEoverridecommandlockouts
\usepackage{cite}
\usepackage{amsmath,amssymb,amsfonts}
\usepackage{algorithmic}
\usepackage{graphicx}
\usepackage{textcomp}
\usepackage{xcolor}
\usepackage{url}
\usepackage{hyperref}

\usepackage{subfig}

\usepackage{xcolor}
\usepackage{eso-pic}
\AddToShipoutPictureBG*{%
  \AtPageUpperLeft{%
    \setlength{\unitlength}{1mm}%
    \put(-9.5,-8){\makebox(\paperwidth,0)[c]{\parbox{0.9\textwidth}{2026 IEEE/ASME International Conference on Advanced Intelligent Mechatronics (AIM), pp. 1-8}}}
  }
}

\AddToShipoutPictureBG*{%
  \AtPageUpperLeft{%
    \setlength{\unitlength}{1mm}%
    \put(-9.5,-13){\makebox(\paperwidth,0)[c]{\parbox{0.9\textwidth}{DOI: \href{https://ieeexplore.ieee.org/document/11658311}{10.1109/AIM65483.2026.11658311}}}}
  }
}

\def\BibTeX{{\rm B\kern-.05em{\sc i\kern-.025em b}\kern-.08em
    T\kern-.1667em\lower.7ex\hbox{E}\kern-.125emX}}
\begin{document}

\title{BinWalker: Development and Field Evaluation \\of a Quadruped Manipulator Platform for Sustainable Litter Collection

\thanks{*Equal contribution. This work is part of the “Technologies for Sustainability” Flagship program of the Istituto Italiano di Tecnologia (IIT). Additional funding by the European Union -
NextGenerationEU and by the Ministry of University and
Research (MUR), National Recovery and Resilience Plan
(NRRP), Mission 4, Component 2, Investment 1.5, project
“RAISE - Robotics and AI for Socio-economic Empowerment” (ECS00000035).}
}

\author{\IEEEauthorblockN{Giulio Turrisi\textsuperscript{*}}
\IEEEauthorblockA{\textit{Dynamic Legged Systems} \\
\textit{Istituto Italiano di Tecnologia (IIT)}\\
Genova, Italy}
\and
\IEEEauthorblockN{Angelo Bratta\textsuperscript{*}}
\IEEEauthorblockA{\textit{Dynamic Legged Systems} \\
\textit{Istituto Italiano di Tecnologia (IIT)}\\
Genova, Italy}
\and
\IEEEauthorblockN{Giovanni Minelli}
\IEEEauthorblockA{\textit{Dynamic Legged Systems} \\
\textit{Istituto Italiano di Tecnologia (IIT)}\\
Genova, Italy}

\and
\hspace{1cm}
\IEEEauthorblockN{Gabriel Fischer Abati}
\IEEEauthorblockA{\hspace{1cm}\textit{Dynamic Legged Systems} \\
\hspace{1cm}\textit{Istituto Italiano di Tecnologia (IIT)}\\
\hspace{1cm}Genova, Italy}

\and
\IEEEauthorblockN{Amir H. Rad}
\IEEEauthorblockA{\textit{Dynamic Legged Systems} \\
\textit{Istituto Italiano di Tecnologia (IIT)}\\
Genova, Italy} 

\and
\IEEEauthorblockN{João Carlos Virgolino Soares
}
\IEEEauthorblockA{\textit{Dynamic Legged Systems} \\
\textit{Istituto Italiano di Tecnologia (IIT)}\\
Genova, Italy}

\and
\hspace{7cm}\IEEEauthorblockN{Claudio Semini}
\IEEEauthorblockA{\hspace{7cm}\textit{Dynamic Legged Systems} \\
\hspace{7cm}\textit{Istituto Italiano di Tecnologia (IIT)}\\
\hspace{7cm}Genova, Italy}
}

\maketitle

\begin{abstract}

Litter pollution represents a growing environmental problem affecting natural and urban ecosystems worldwide. Waste discarded in public spaces often accumulates in areas that are difficult to access, such as uneven terrains, coastal environments, parks, and roadside vegetation. Over time, these materials degrade and release harmful substances, including toxic chemicals and microplastics, which can contaminate soil and water and pose serious threats to wildlife and human health. Despite increasing awareness of the problem, litter collection is still largely performed manually by human operators, making large-scale cleanup operations labor-intensive, time-consuming, and costly. Robotic solutions have the potential to support and partially automate environmental cleanup tasks. In this work, we present a quadruped robotic system designed for autonomous litter collection in challenging outdoor scenarios. The robot combines the mobility advantages of legged locomotion with a manipulation system consisting of a robotic arm and an onboard litter container. This configuration enables the robot to detect, grasp, and store litter items while navigating through uneven terrains. The proposed system aims to demonstrate the feasibility of integrating perception, locomotion, and manipulation on a legged robotic platform for environmental cleanup tasks. Experimental evaluations conducted in outdoor scenarios highlight the effectiveness of the approach and its potential for assisting large-scale litter removal operations in environments that are difficult to reach with traditional robotic platforms. 
The code associated with this work can be found at:
\url{https://github.com/iit-DLSLab/trash-collection-isaaclab}.
\end{abstract}


\begin{figure}
\centering

\includegraphics[width=0.66\linewidth]{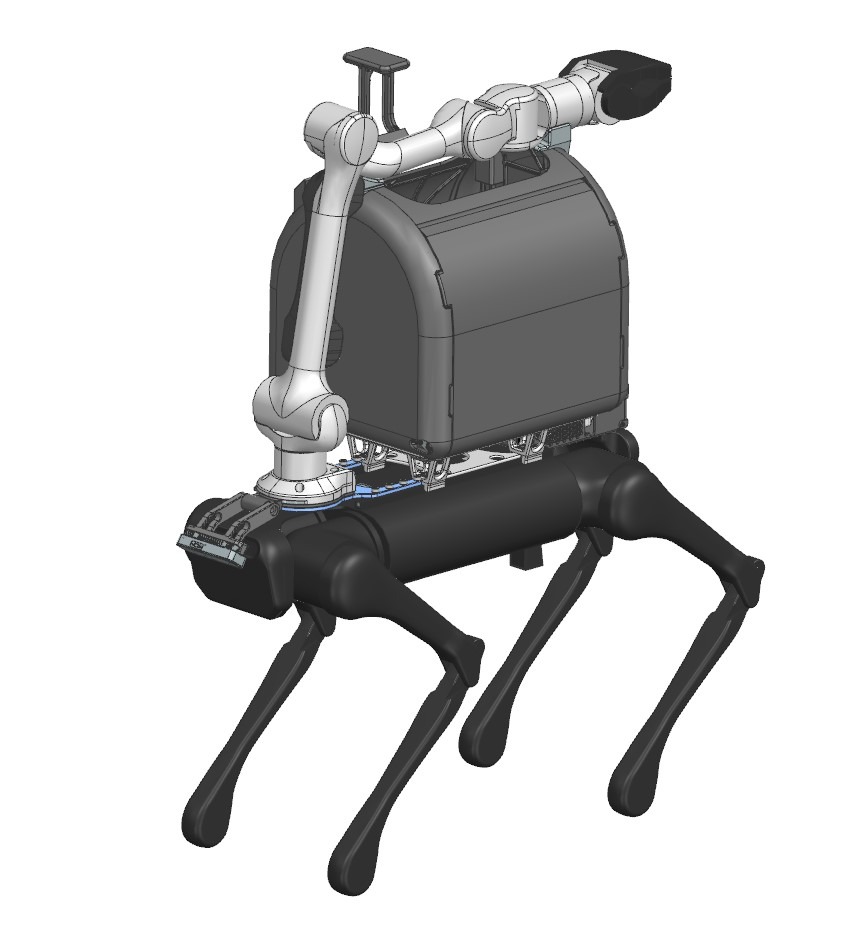}
\includegraphics[width=0.54\linewidth]{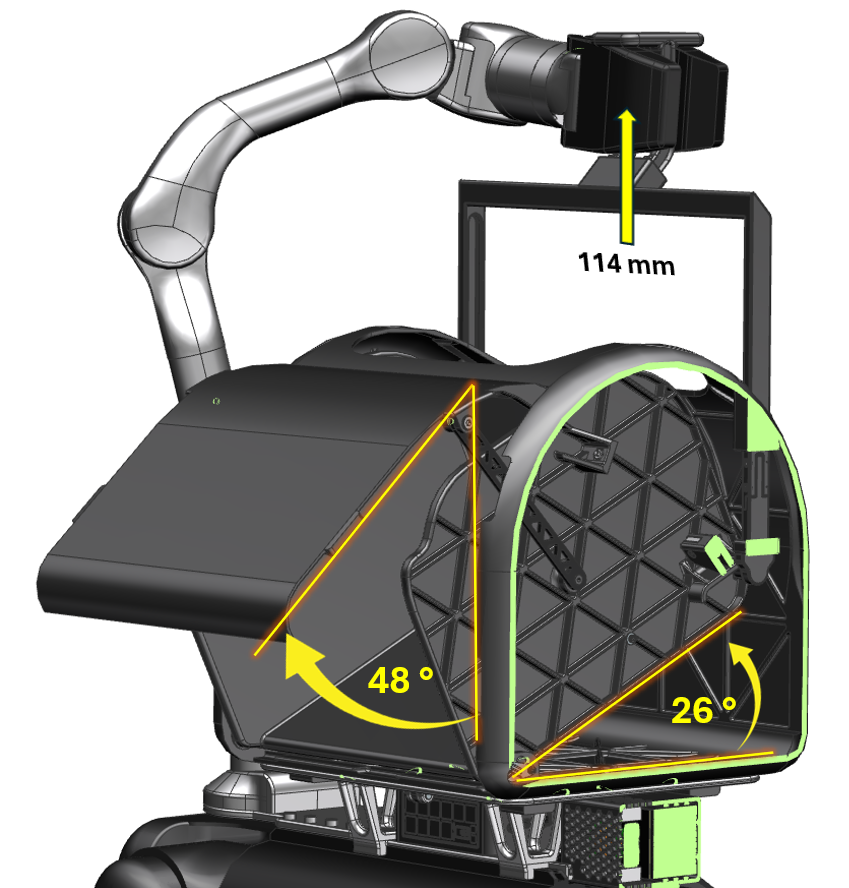}\\[0.6em]

\caption{
Prototype based on the Aliengo robot. Top: Isometric view of the robot highlighting the components mounted on the top of the trunk, including the frontal camera, the Z1 arm, and the litter container. Bottom: detailed view of the container opening mechanism used to unload the collected litter.
}
\label{fig:cads}
\end{figure}

\begin{figure*}[ht]
\centering
\includegraphics[width=0.99\linewidth]{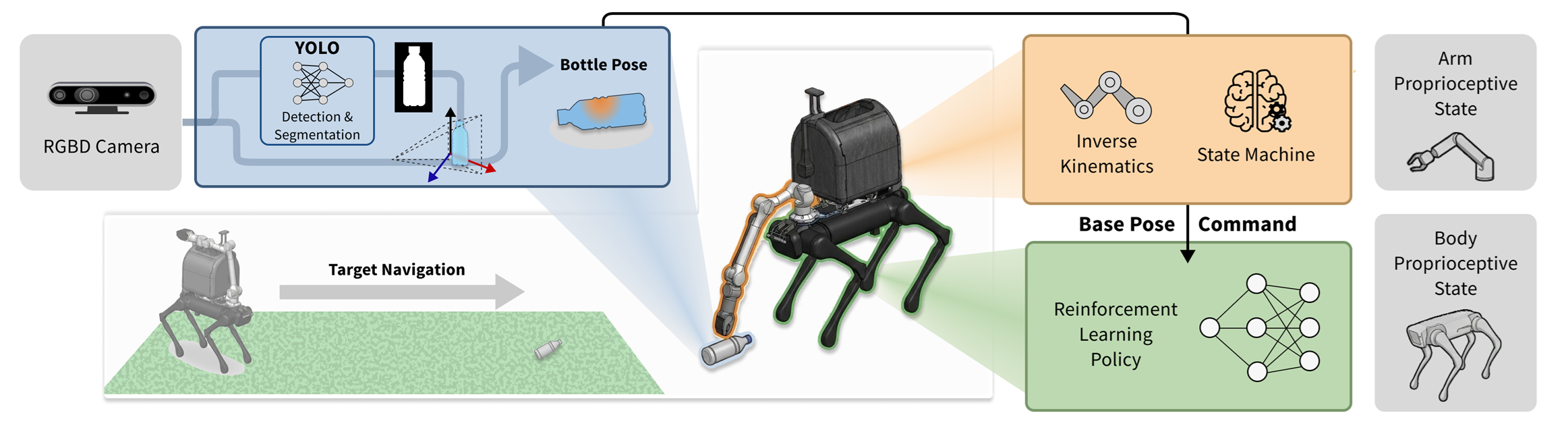}
\caption{Block scheme of the proposed approach.}
\label{fig:block_scheme}
\end{figure*}

\section{Introduction}
Litter pollution represents a growing environmental challenge affecting both urban and natural ecosystems \cite{UNEP}. Waste discarded in public areas often accumulates in locations such as parks, coastal regions, and roadside environments, where it may persist for long periods and progressively degrade into harmful substances, including toxic chemicals and microplastics. These pollutants can contaminate the soil and water and pose a serious threat to wildlife and human health. Despite increasing awareness of this issue, litter removal is still largely performed manually by humans \cite{Ocean}, making large-scale cleanup operations labor-intensive, costly, and difficult to sustain over time.

Robotic systems have recently emerged as a promising solution to support environmental cleanup tasks. Several autonomous platforms have been proposed for litter detection and collection in outdoor environments \cite{mittal2016spotgarbage}. Most of these systems are based on wheeled or tracked robots operating in structured or semi-structured settings such as sidewalks, beaches, or urban pavements \cite{oatcr2021}. Although such platforms can be effective in relatively flat terrains, their mobility becomes significantly limited in environments characterized by uneven ground, vegetation, obstacles, or narrow passages.

Legged robots offer an attractive alternative for these scenarios due to their superior mobility and terrain adaptability \cite{vero}. In recent years, quadruped robots have demonstrated remarkable capabilities in traversing rough terrain, climbing obstacles, and navigating complex outdoor environments \cite{hwangbo2019learning, lee2020perceptive, rudin2022minutes}. These characteristics make them particularly suitable for environmental monitoring and intervention tasks in areas that are difficult to access using traditional mobile robots.

Alongside mobility, perception, and manipulation capabilities are essential for autonomous litter collection. Recent advances in deep learning have significantly improved the performance of vision-based object detection systems, enabling reliable identification of litter in cluttered outdoor environments \cite{ren2015faster,redmon2016yolo, kirillov2023sam}. Convolutional neural networks have been successfully applied to detect various categories of litter, including plastic debris, bottles, and cigarette butts, providing a robust perception pipeline for robotic cleanup applications \cite{taco2020, mittal2016spotgarbage}. However, integrating perception with efficient planning and manipulation remains a challenging problem, especially when operating on mobile robotic platforms in dynamic outdoor environments.

Regarding mobile manipulation,  different works have studied the problem of loco-manipulation \cite{mani_1},\cite{mani_2}, \cite{mani_3} under a non-specific setting. Here, instead, we tailor this problem in the context of litter-collection, with the goal of studying the challenges that are inherently related to this task.

In this work, we present a quadruped robotic system designed for autonomous litter collection in challenging outdoor environments. The proposed platform combines the terrain adaptability of legged locomotion with a robotic arm and an onboard litter container, enabling the robot to detect, grasp, and store litter items while navigating through irregular terrains. Litter detection is performed using a convolutional neural network trained to identify litter in outdoor scenes. Collection is then decoupled by a hierarchical approach, separating arm and base pose optimization which relies on standard Inverse Kinematics (IK), from locomotion, which instead is developed by adopting a Reinforcement Learning (RL) algorithm to reach robust walking over challenging terrains.

\subsection{Contribution}

The main contributions of this work can be summarized as follows:
\begin{itemize}

\item A mobile robotic platform for autonomous litter collection that integrates quadruped locomotion, manipulation, perception, and onboard litter storage.

\item An experimental evaluation demonstrating the feasibility of autonomous litter collection using a legged robotic platform.
\end{itemize}

Furthermore, this manuscript discusses practical insights, highlighting what works well and identifying aspects that should be improved to enhance the performance of autonomous robots in this task. Finally, we release the code associated with this work with the goal of providing the community with a baseline upon which their results can be compared or further research can be built.

\subsection{Outline}
The paper is organized as follows: Sec. \ref{sec:hardware} gives an overview
of our litter collection prototype, whereas Sec. \ref{sec:method} describes the
proposed approach delving into the components of locomotion, manipulation, and vision. In Section \ref{sec:results}, we report on the application of the proposed prototype in an outdoor experimental scenario, drawing discussion of current limitations and possible future work directions.

\section{Litter Collection Hardware}

\label{sec:hardware}
For this work, the commercial quadruped robot Unitree Aliengo \footnote{Unitree Aliengo: \url{https://www.unitree.com/mobile/aliengo/}} has been adapted by integrating a robotic manipulation and storage system to enable autonomous litter collection (see Fig. \ref{fig:cads}-top). A 6-DoF lightweight manipulator, the Unitree Z1 robotic arm \footnote{Unitree Z1 Pro: \url{https://www.unitree.com/z1}}, is mounted in the frontal section of the robot trunk through a custom aluminum support frame. This configuration, coupled with the position of the stereo camera Realsense D435 under the base of the arm, allows the manipulator to operate within a large frontal workspace.

On top of the trunk of the robot, right behind the manipulator arm, we placed a cabinet acting as a bin container. The cabinet consists of a rigid ABS outer body (310 × 360 × 445.5 mm, 2.6 kg) with an open top for litter entry, housing an internal rotating basket held by magnets to dampen vibrations during locomotion. A hinged door at the base is mechanically linked to the basket via two connecting links, so when the robotic arm lifts the detachable handle 114 mm upward, the basket rotates 26° and the door swings open 48° simultaneously to discharge the litter. After unloading, gravity resets the entire mechanism to its original position. An exploded view of the cabinet can be found in Fig. \ref{fig:cads}-bottom, while a detailed description in \cite{amir}.

\section{Litter Collection Methodology}
\label{sec:method}
In this section, we describe the main components of our collection approach. These include Locomotion, which enables the robot to traverse different scenarios; Manipulation, which controls the movement of the manipulator arm and the robot's pose and is activated once the system reaches the proximity of a litter item to be collected; and Vision, which provides the Locomotion and Manipulation modules with information about the presence and location of nearby litter items.

These components operate in coordination to achieve a successful grasp. Their interaction and overall workflow are illustrated in the block diagram shown in Fig. \ref{fig:block_scheme}.

\subsection{Locomotion}
\label{subsec:locomotion}
    The locomotion controller finds the best joint leg torques in order to move the robot in space. Different methods exist in the literature to perform such locomotion, such as model-based control \cite{mcts_modelbased} and RL \cite{cheng2023}. Although both methods are able to perform agile walking, RL emerges for robustness, being able to not rely on fixed gait sequences. Still, loco-manipulation entirely using RL presents different challenges, such as a lack of precision and difficulty in training. For this reason, in this work, we decoupled locomotion and manipulation, similar to \cite{mani_1}. We develop an RL-based policy that optimizes leg movements, which can follow and realize arbitrary base velocities and base poses; the latter serve to increase the grasping workspace, that is performed by a separate arm controller detailed in Sec. \ref{subsec:manipulation}.

The locomotion policy observations depend primarily on proprioceptive information, such as:

\begin{equation}
    \mathbf{o_\mathrm{p}} = (\mathbf{v}, \mathbf{w}, 
    \mathbf{q_\mathrm{leg}}, \dot{\mathbf{q}}_\mathrm{leg}, \mathbf{q_{\mathrm{arm}}})
    \label{eq:observation_proprio}
\end{equation}

\noindent where $\mathbf{v}, \mathbf{w} \in \mathcal{R}^3$ are, respectively, the base linear and angular velocity of the robot, provided by a state estimator~\cite{muse}; $\mathbf{q}$ and $\dot{\mathbf{q}}$  are the joints' position and velocities. 
In particular, $\mathbf{q}_\mathrm{leg}$ and $\dot{\mathbf{q}}_\mathrm{leg} \in \mathcal{R}^{12}$ represent the variable for the quadruped, 
while $\mathbf{q}_\mathrm{arm} \in \mathcal{R}^6$ is associated with the manipulator. During training, given the separation between the locomotion and manipulation controllers, we randomly assign the initial joint positions of the arm $\mathbf{q_{\mathrm{arm}}}$ and maintain them fixed, to robustify the locomotion policy during grasping. For this, no $\mathbf{\dot{q}_{\mathrm{arm}}}$ are added inside the observation.
Starting from the proprioceptive information $\mathbf{o_p}$, we then append the reference values to be followed by the policy, such as:

\begin{equation}
    \mathbf{o_\mathrm{r}} = (
    \mathbf{v^\mathrm{des}}, \mathbf{w^\mathrm{des}}, 
    h^\mathrm{des}, \theta^\mathrm{des})
\label{eq:observation_ref}
\end{equation}

\noindent where, $h$ and $\theta \in \mathcal{R}$ are the desired height and pitch of the base. As stated previously, optimizing the base pose as well augments the reachable workspace of the arm, reducing the drawbacks of having two separate controllers, which can now communicate with these auxiliary variables. Furthermore, the final policy input is represented by the stack of 5 past observations \eqref{eq:observation_proprio}-\eqref{eq:observation_ref} in order to make the policy adaptive.

The output of the policy, instead, is the commonly used desired joint positions $\mathbf{q}^\mathrm{des}_{\mathrm{leg}} \in \mathcal{R}^{12}$ which are tracked by a simple \textit{proportional-derivative} low-level controller.

The rewards function used in this work can be found in Table \ref{tab:rewards}. In the table, we define with $\tilde{\mathbf{p}}$ the distance between the xy component of the foot wrt the associated robot's hip in the horizontal frame; with $(\cdot)^\textrm{terrain}$ the quantities that we obtain from the simulator, representing the for example the terrain orientation $\theta^\textrm{terrain}$ and the height of the terrain $h^\textrm{terrain}$; finally with $c^\textrm{des}, c$, the contacts suggestion coming from a contact generator with step frequency equal to 1.4 and duty factor equal to 0.6, and the ground truth foot contact coming from the simulators. Notably, we employ a periodic clock signal to generate contact and provide the policy with a continuous reward signal to optimize its actions. Even if this choice could seem limiting, this is merely a suggestion to the policy, which is still able to break and make contact with the environment at need. Furthermore, in the table, we remove the subscript leg to the variable $\mathbf{q}$ for clarity.

\begin{table}[t]
\vspace{0.3cm}
\renewcommand{\arraystretch}{1.0}
\centering
\caption{Locomotion policy rewards}
\begin{tabular}{c|c}
\hline
\textbf{Type} & \textbf{Expression} \\ 
\hline
base linear velocity & $\exp(-(\mathbf{v}^\textrm{des} - \mathbf{v})^2/0.25)$\\ 
base angular velocity & $-(\mathbf{w}^\textrm{des} - \mathbf{w})^2$ \\
base orientation & $-(\theta^\textrm{terrain} + \theta^\textrm{des} - \theta)^2$\\
base height & $\exp(-(h^\textrm{des} - h)^2/0.01)$\\
joints torque & $-\boldsymbol{\tau}^2$\\
joints acceleration & $-\ddot{\mathbf{q}}^2$\\
joints energy & $-|\dot{\mathbf{q}} \cdot \boldsymbol{\tau}|$\\
undesired contact & $-1 \textrm{ for each base collision}$\\
action rate & $-(\mathbf{q}^\textrm{des}_k - \mathbf{q}^\textrm{des}_{k-1})^2$\\
action smoothness & $-(\mathbf{q}^\textrm{des}_k - 2\mathbf{q}^\textrm{des}_{k-1} - \mathbf{q}^\textrm{des}_{k-2})^2$\\
feet contact suggestion & $\sum_{4} -(c^\textrm{des} - c)$\\
feet height clearance &$ \sum_{4} \exp(-(\mathrm{p}^\textrm{des}_\textrm{foot,z} + h^\textrm{terrain} - \mathrm{p}_\textrm{foot,z} )^2/0.01)$\\
feet to hips position & $-(\tilde{\mathbf{p}}^\textrm{ref}_\textrm{foot,xy} - \tilde{\mathbf{p}}_\textrm{foot,xy})^2$\\

\end{tabular}
\label{tab:rewards}
\end{table}

\subsection{Manipulation}
\label{subsec:manipulation}
As described in the previous section, in this work, we opted for a separation between locomotion and manipulation controllers due to the inherent complexity of a whole-body architecture. 

The manipulation controller is composed of two different layers that are coordinated by a state-machine to reach a successful grasp: 1) a layer composed of a set of predefined primitives in joint spaces; 2) an inverse kinematic solver that works in Cartesian space. 

Starting from the first one, some motions, such as putting the object inside the litter cabinet, or emptying it when needed, can be considered repetitive by nature. Every time one of the two task needs to be accomplished in fact, a preplanned joint position can be reached. For this reason, predefined keypoints in the joint space are retrieved and simply tracked by a low-level controller. The primitives implemented in this work are six: \textit{litter-loading}, which brings the collected object on top of the cabinet; \textit{box-reaching}, which makes the arm reach the handle of the litter box; \textit{box-opening}, which makes the arm lift the handle of the litter box;  \textit{rest-pose}, which is the standard configuration that the arms has during walking;  \textit{open-gripper}, \textit{close-gripper}, which handle the gripper motions.

The Inverse Kinematics layer, instead, generates the desired joint positions to reach, with the end-effector, the object to grasp. For this, we employed Mink \cite{mink}, an open-source project built on top of MuJoCo \cite{todorov2012mujoco} that solves differential kinematics by weighted tasks. We start from the main error function

\begin{equation}
    \mathbf{e}_{\mathrm{ee}}\mathbf{(q_\textrm{ik})} = \mathbf{p}^{\mathrm{des}}_{\mathrm{ee}} - \mathbf{p}_{\mathrm{ee}}(\mathbf{q}_\textrm{ik})
\end{equation}

which defined the error between the reference pose to be reached - which, in our case, corresponds to the grasp position coming from the vision module and the current pose of the end-effector of the arm. We then add an additional task for inputs regularization:

\begin{equation}
\mathbf{e}_{\mathrm{reg}}\mathbf{(q_\mathrm{ik})} = \mathbf{q}_\mathrm{ik}^{\mathrm{des}} - \mathbf{q}_\mathrm{ik}
\end{equation}

which can prevent the solver from generating unnecessary high joint displacement maneuvers.

Mink computes the inputs $\dot{\mathbf{q}}_\textrm{ik}$ by solving a quadratic program in the form of 

\begin{equation}
\begin{aligned}
\min_{\dot{\mathbf{q_\textrm{ik}}}} \quad & \sum_{\text{task } e} \left\| \mathbf{J}_\mathbf{e} \mathbf{(q_\textrm{ik})}\mathbf{\dot{\mathbf{q}}_\textrm{ik}} + \mathbf{\alpha} \mathbf{e(q_\textrm{ik})} \right\|_{W_e}^{2} \\
\text{subject to} \quad & \mathbf{h}_{\min}(\mathbf{q_\textrm{ik}}) \le \mathbf{h} \le \mathbf{h}_{\max}(\mathbf{q_\textrm{ik}})
\end{aligned}
\end{equation}

where $\mathbf{J}_e$ is the task jacobian, $\mathbf{\alpha}$ the weighting scalars for giving different priority to different tasks, and $\mathbf{h}$ collision constraints.

\begin{figure}
\centering
\vspace{0.1cm}
\includegraphics[width=0.99\linewidth]{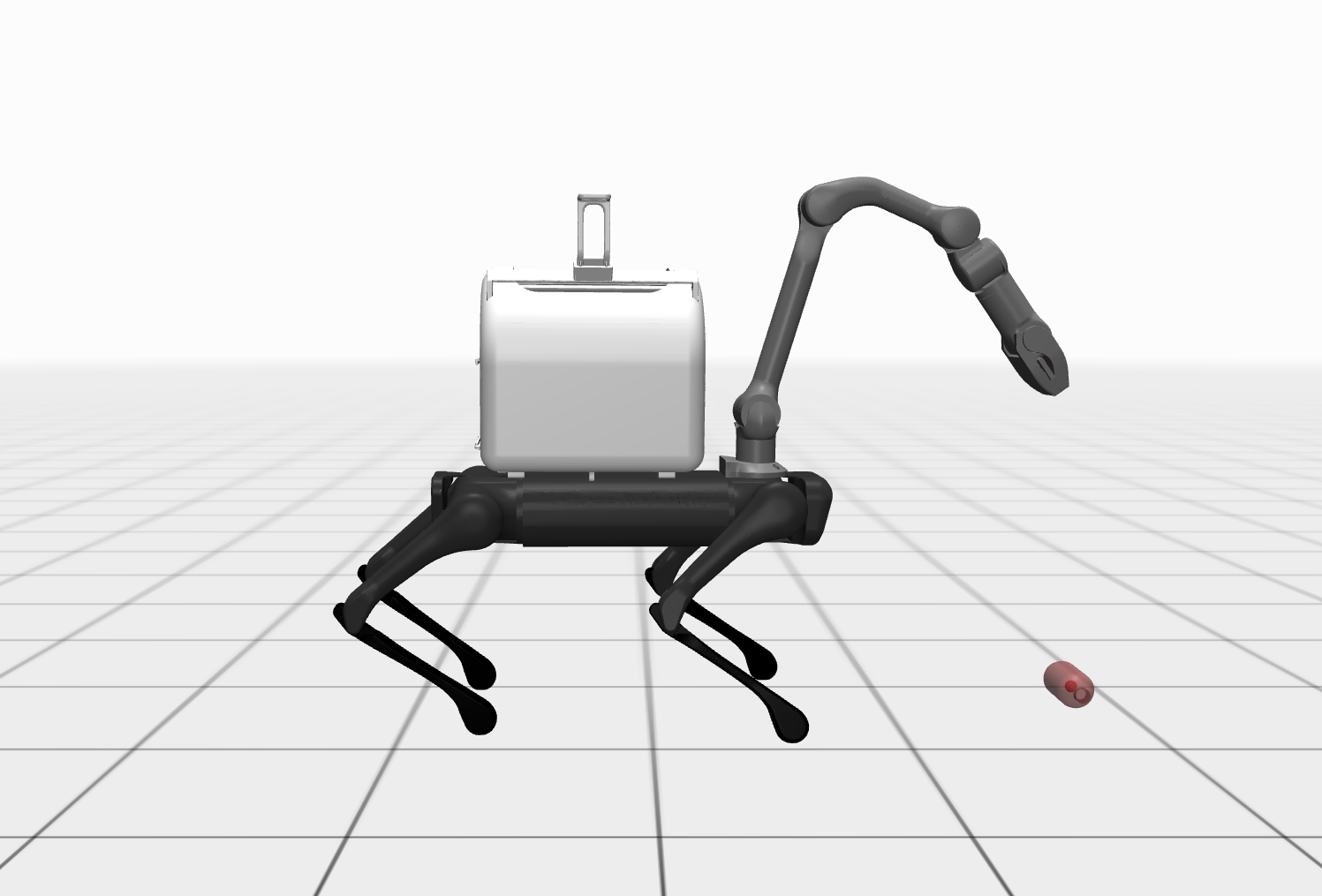}\\[0.3em]
\includegraphics[width=0.99\linewidth]{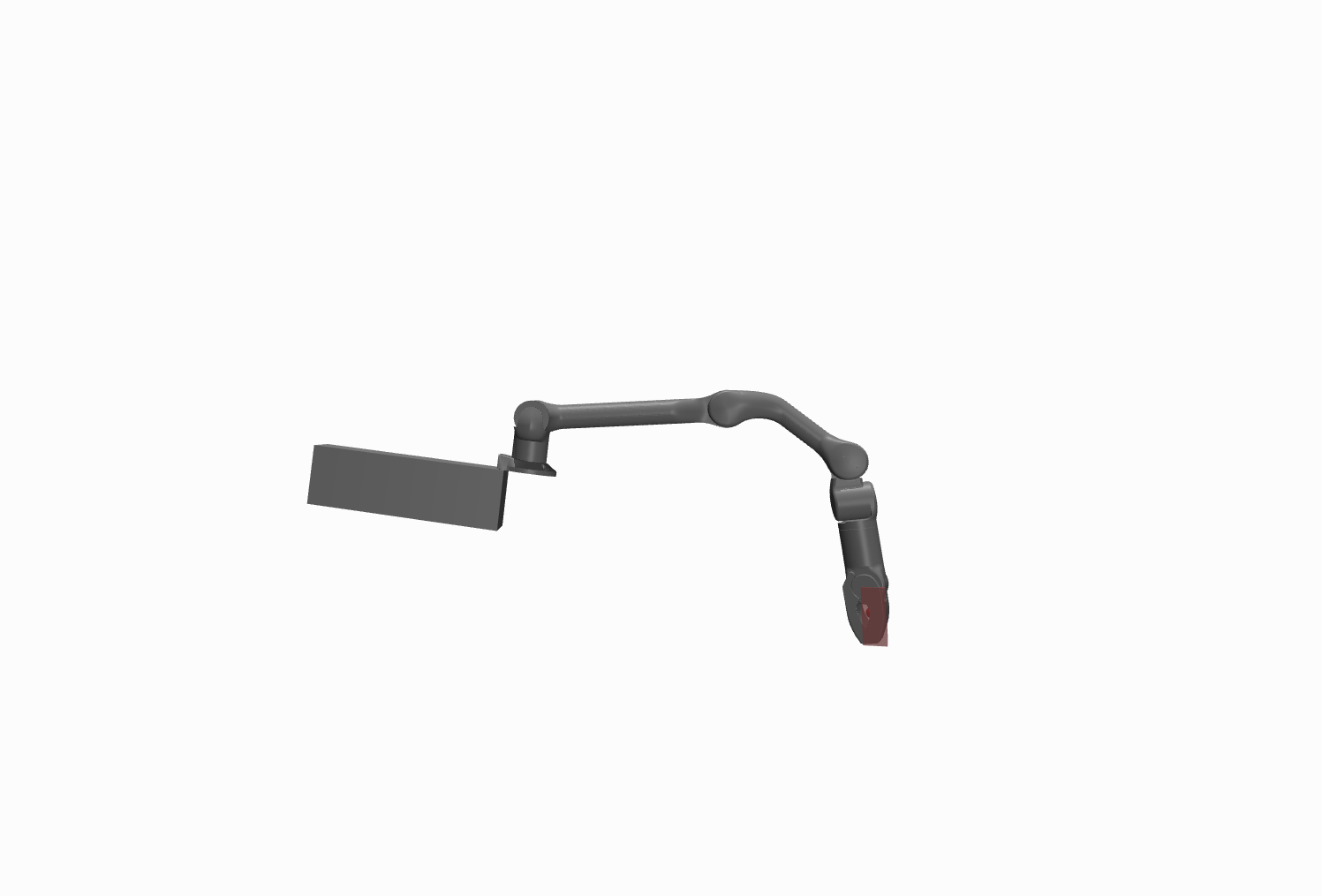}

\caption{
Snapshot of a collection procedure performed in simulation (top). The IK layer, which provides the desired grasping configuration, acts on a simplified system that neglects legs, optimizing arm joints' positions and base pose (height and pitch) in order to reach the desired end-effector pose represented in the figure by a red box (bottom).
}
\label{fig:ik}
\end{figure}

In our case, we employ Mink over a simplified robot model. As stated in Sec. \ref{subsec:locomotion}, the locomotion policy can modify at need the height and the pitch of the robot. This serves to enlarge the workspace of the arm. For this reason, the IK layer can optimize not only arm joints position, but $h^\mathrm{ref}, \theta^\mathrm{ref}$ are considered as additional input variables, obtaining $\mathbf{q}_\mathrm{ik} = (\mathbf{q}_\mathrm{leg}^\mathrm{des}, h^\mathrm{des}, \theta^\mathrm{des})$. In order to prioritize the movement of the arm, and only move the base when needed, a bigger value of $\alpha$ is associated to the task that regularizes the body pose. Fig. \ref{fig:ik} depicts our hierarchical control in action inside our simulation environment.

Finally, a state machine is in charge of switching between the predefined primitives, concatenating them in order to perform the desired task (e.g. \textit{open-gripper} $\to$ \textit{box-reaching} $\to$ \textit{close-gripper} $\to$ \textit{box-opening}). Furthermore, it handles the exchange between the primitives and the IK layers. This is simply done by checking if a detection is received by the vision module (Sec. \ref{subsec:vision}), and activating/deactivating the IK at need.

\subsection{Vision}
\label{subsec:vision}

The objective of the orientation estimation module is to estimate the 3D position of a bottle and the direction of its longitudinal axis from RGB-D observations. This information is required to align the robotic gripper with the object prior to grasping. The method takes as input an RGB image, a depth map aligned with the RGB frame, and the camera intrinsic parameters. The output is a 3D point representing the bottle position and a unit vector describing the direction of its principal axis in the camera coordinate frame.

\begin{figure}[t]
\centering
\vspace{0.1cm}
\includegraphics[width=0.47\linewidth]{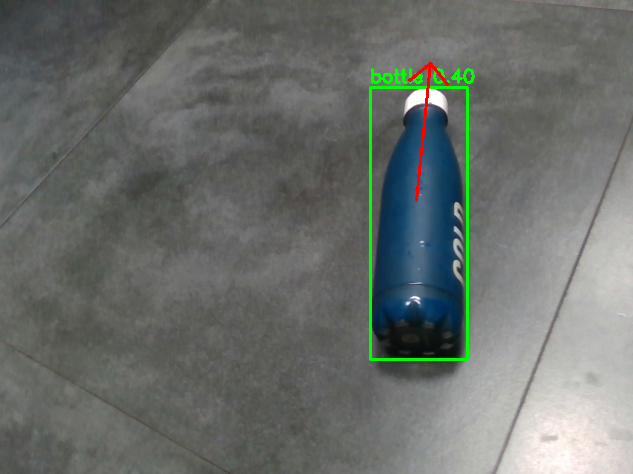}
\includegraphics[width=0.47\linewidth]{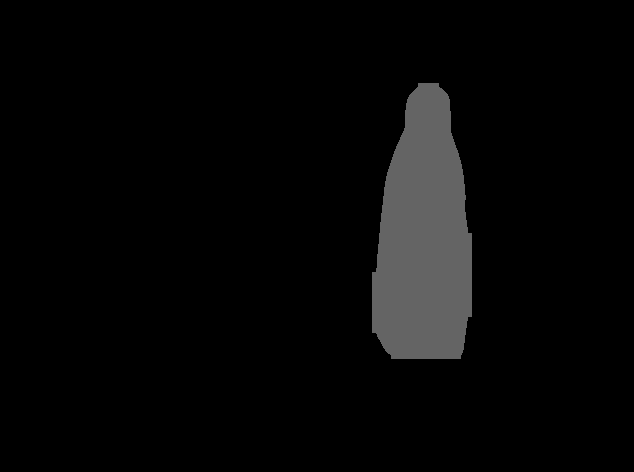}

\caption{The segmentation mask (right) is used to compute the principal axis of the bottle via PCA. The reconstructed 3D axis is projected back onto the RGB image for visualization (left).}
\label{fig:three_images}
\end{figure}

In this work, we focus on plastic bottles, which represent a common form of urban litter. Bottle detection and segmentation are performed using a YOLO-based instance segmentation network~\cite{jocher2023ultralytics}, which directly predicts pixel-wise masks for detected objects. The network identifies instances belonging to the bottle class and produces a binary segmentation mask representing the object silhouette, shown in Fig.~\ref{fig:three_images} (right). The segmentation mask is further refined by selecting the largest connected component and applying morphological filtering to remove small artifacts and improve shape consistency.

Given the binary mask, the dominant orientation of the bottle is estimated in the image plane using Principal Component Analysis (PCA). Let $\mathbf{p}_i = (u_i, v_i)$ denote the pixel coordinates belonging to the mask. The centroid of the mask is computed as

\begin{equation}
\bar{\mathbf{p}} = \frac{1}{N} \sum_{i=1}^{N} \mathbf{p}_i .
\end{equation}
The covariance matrix of the mask pixels is then calculated as
\begin{equation}
\mathbf{C} = \frac{1}{N-1} \sum_{i=1}^{N} (\mathbf{p}_i - \bar{\mathbf{p}})(\mathbf{p}_i - \bar{\mathbf{p}})^{T}.
\end{equation}

The eigenvector associated with the largest eigenvalue of $\mathbf{C}$ corresponds to the direction of maximum variance and provides an estimate of the bottle’s principal axis in the image plane.

To recover the orientation in 3D space, depth values are sampled at representative points along the estimated 2D axis and back-projected using the camera intrinsic matrix
\begin{equation}
\mathbf{K} =
\begin{bmatrix}
f_x & 0 & c_x \\
0 & f_y & c_y \\
0 & 0 & 1
\end{bmatrix}.
\end{equation}

Given a pixel $(u,v)$ with depth $Z$, the corresponding 3D point $\mathbf{P}=(X,Y,Z)$ in the camera frame is obtained as
\begin{equation}
X = \frac{(u - c_x)Z}{f_x}, \qquad
Y = \frac{(v - c_y)Z}{f_y}, \qquad
Z = Z.
\end{equation}

Two reconstructed points $\mathbf{P}_1$ and $\mathbf{P}_2$ along the axis are used to estimate the bottle direction
\begin{equation}
\mathbf{a} = \frac{\mathbf{P}_2 - \mathbf{P}_1}{\|\mathbf{P}_2 - \mathbf{P}_1\|}.
\end{equation}
The bottle position is estimated as the midpoint
\begin{equation}
\mathbf{c} = \frac{\mathbf{P}_1 + \mathbf{P}_2}{2}.
\end{equation}

Finally, temporal stabilization is applied to improve robustness across consecutive frames. Since the principal axis is directionally ambiguous, a sign consistency constraint is enforced using the orientation estimated in the previous frame. Additionally, exponential smoothing is applied to reduce jitter caused by segmentation noise and depth measurement uncertainty. The resulting pose estimate provides a stable 3D position and orientation that can be directly used by the robotic manipulator to align the gripper with the bottle’s main axis during grasping. For this, we then map the detection in the base frame of the robot and pass it to the IK module to perform the final grasp.

Although the pose estimation method is described for plastic bottles, the approach can be extended to other types of litter commonly encountered in outdoor environments. The perception pipeline relies on instance segmentation to isolate objects, followed by geometric analysis of the resulting mask. Additional litter categories can be incorporated by extending or retraining the segmentation network to detect other objects such as aluminum cans, disposable cups, or small plastic containers. For elongated objects, such as bottles or cans, the proposed principal axis estimation can be used for gripper alignment, while for objects with less defined geometry simpler grasping strategies based on the object centroid can be adopted.

\section{Results}
\label{sec:results}

\begin{figure*}
\centering

\includegraphics[width=0.328\linewidth]{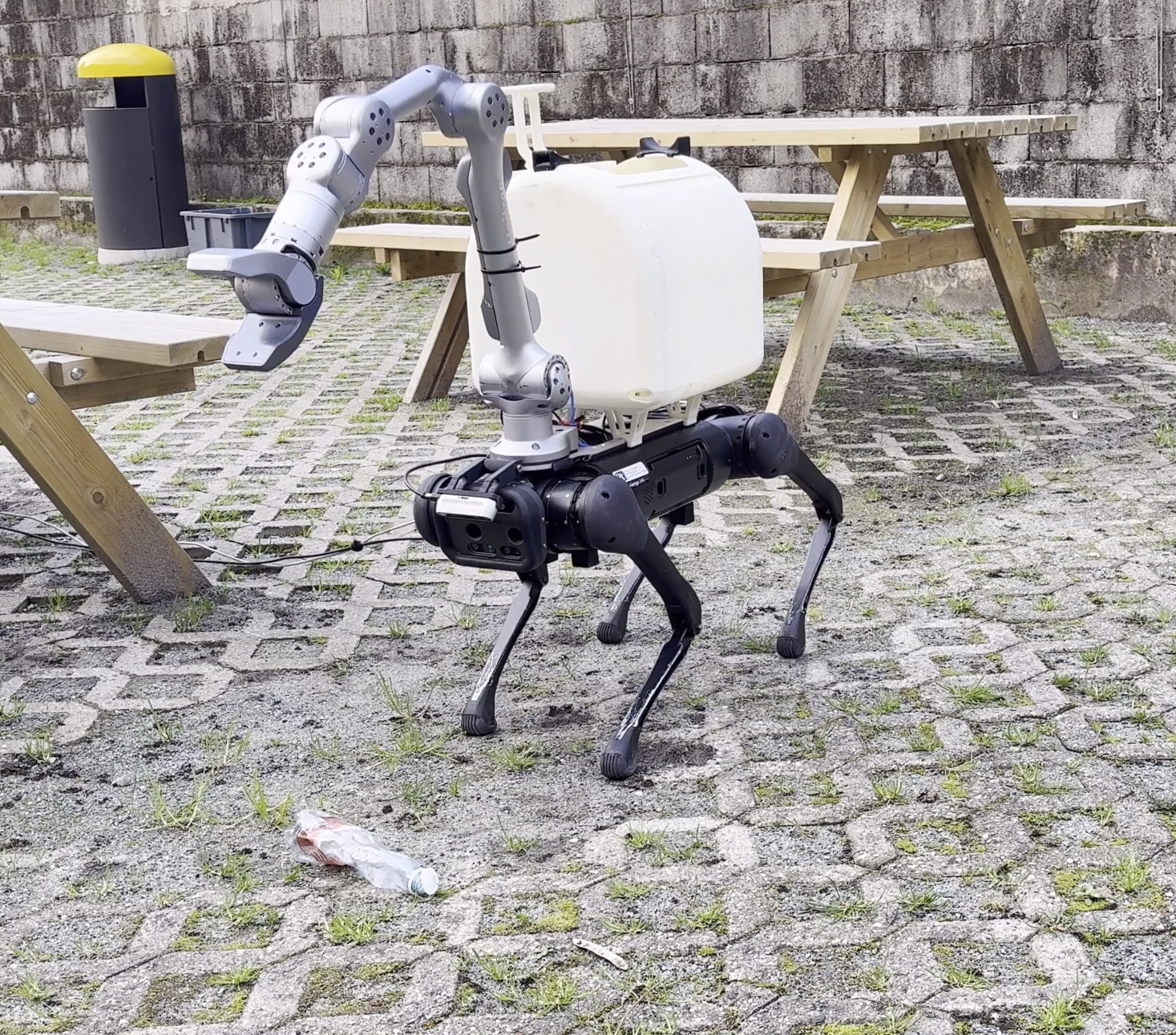}
\includegraphics[width=0.328\linewidth]{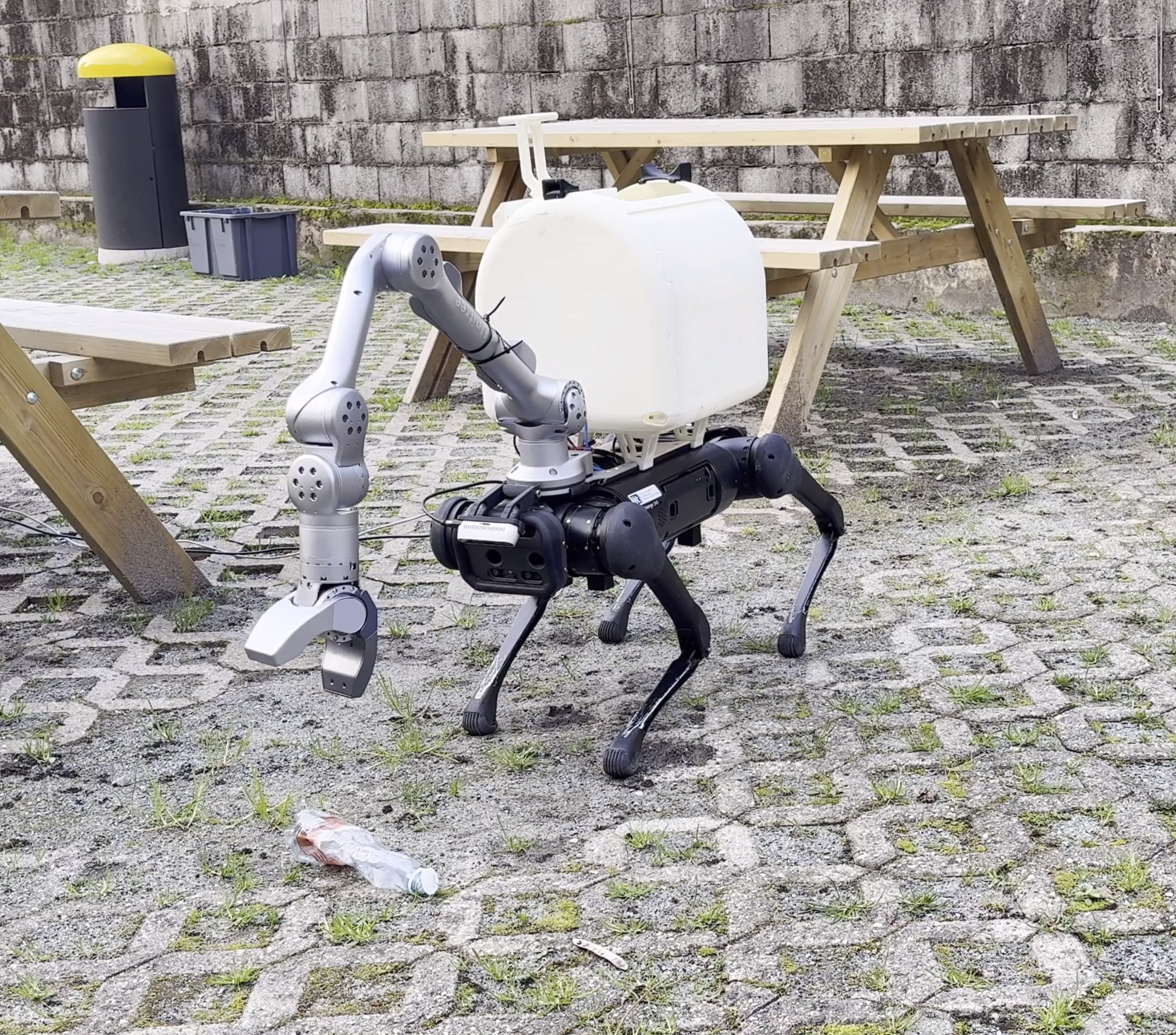}
\includegraphics[width=0.328\linewidth]{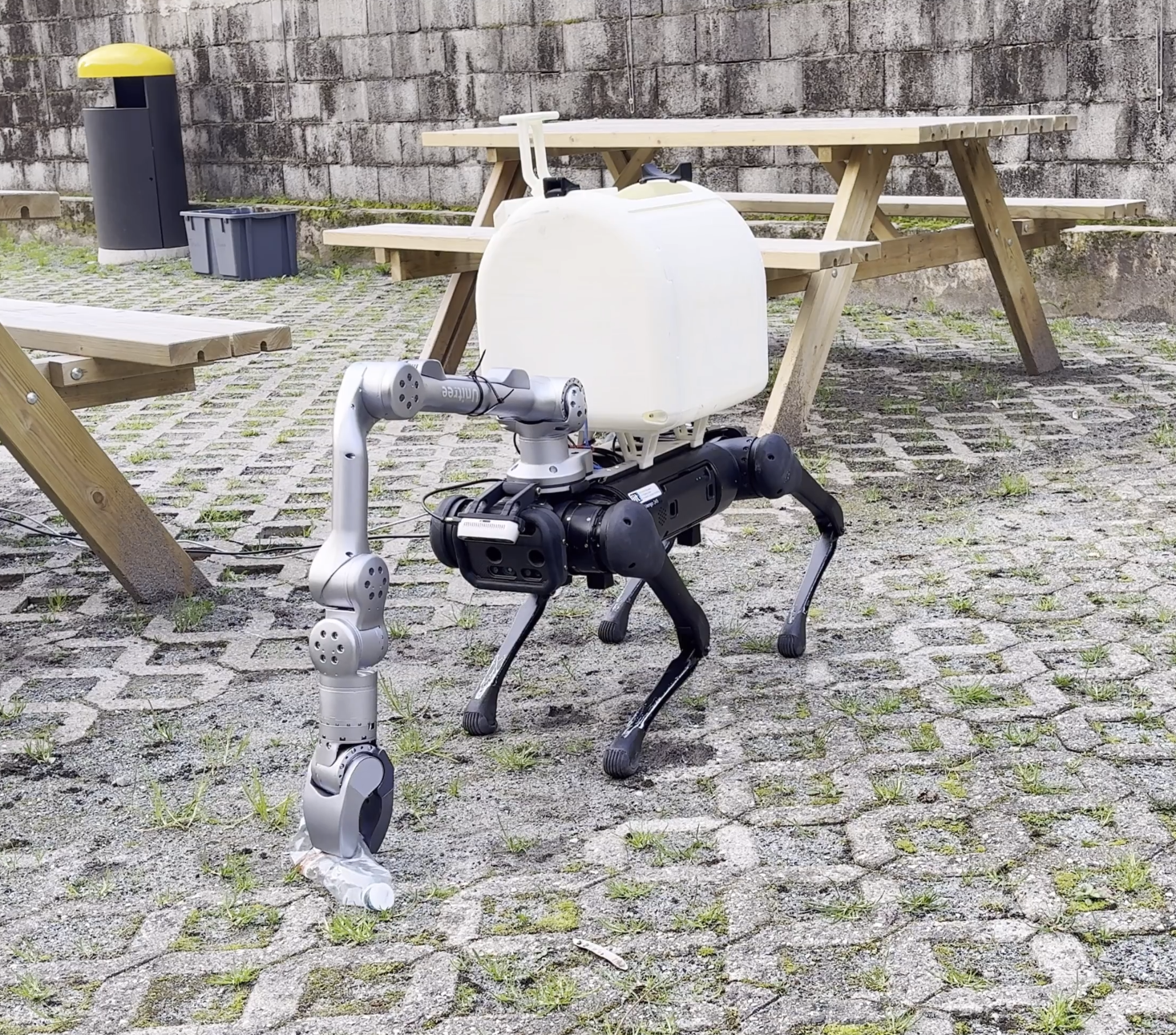} \\[0.3em]

\includegraphics[width=0.328\linewidth]{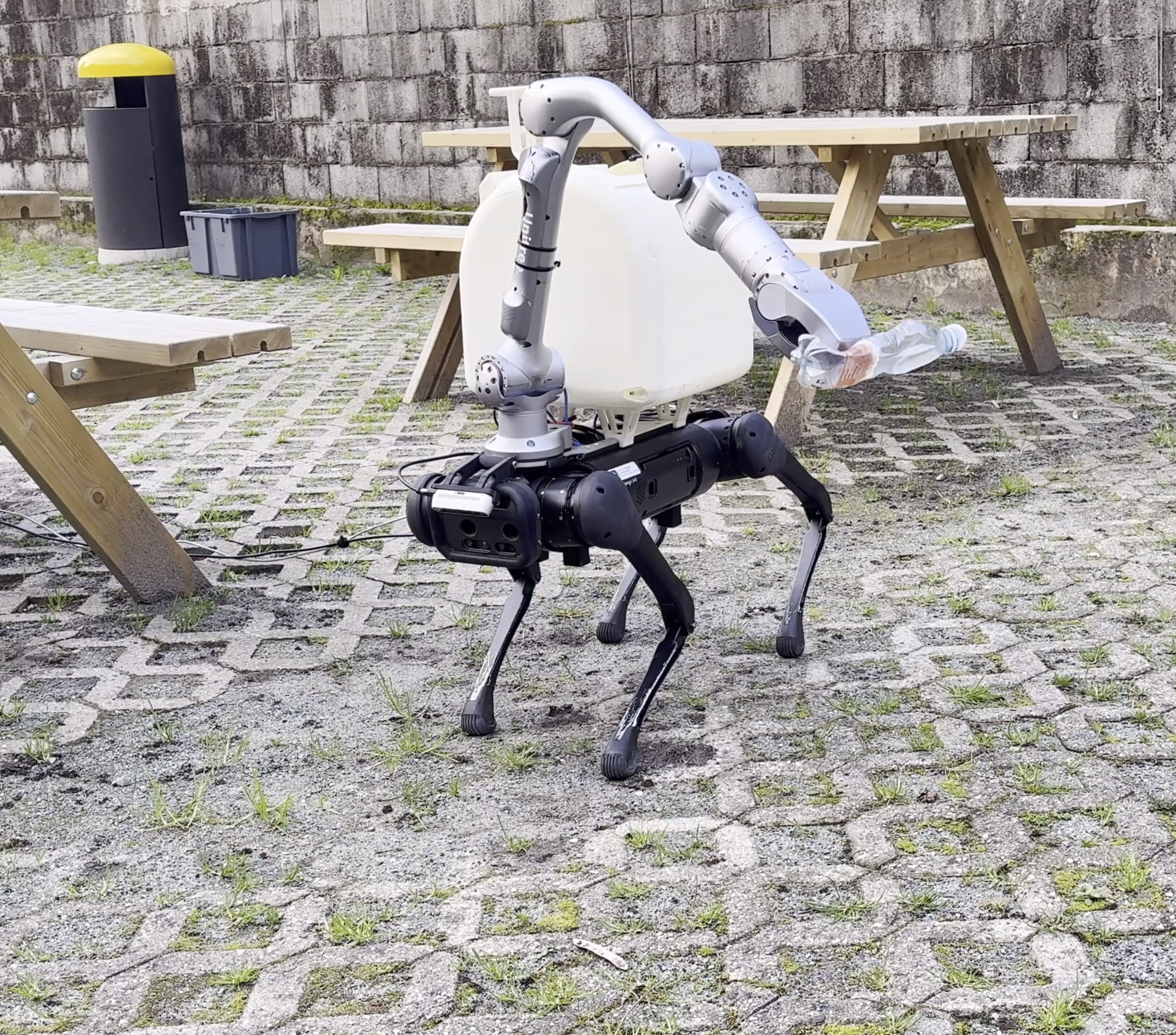}
\includegraphics[width=0.328\linewidth]{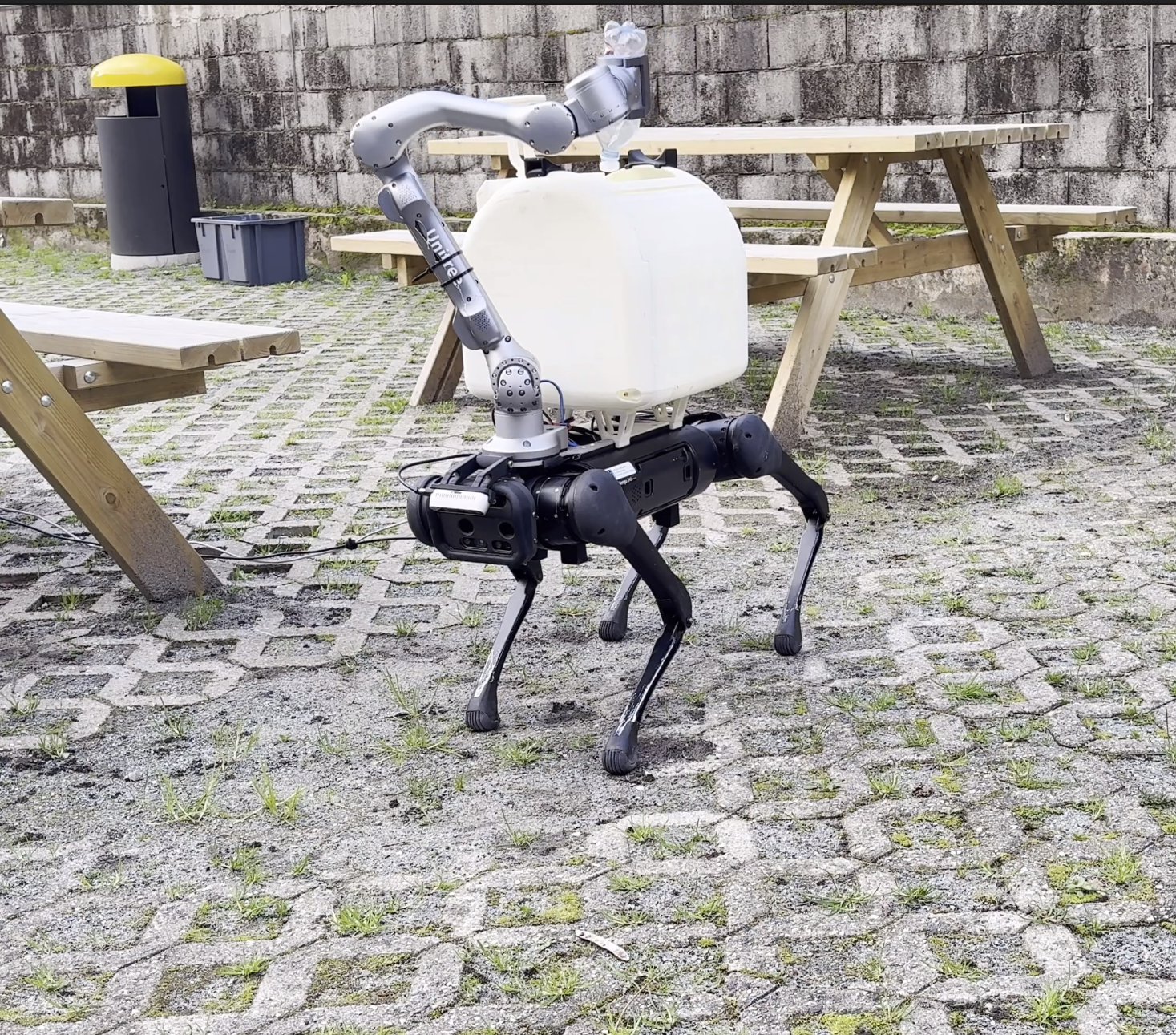}
\includegraphics[width=0.328\linewidth]{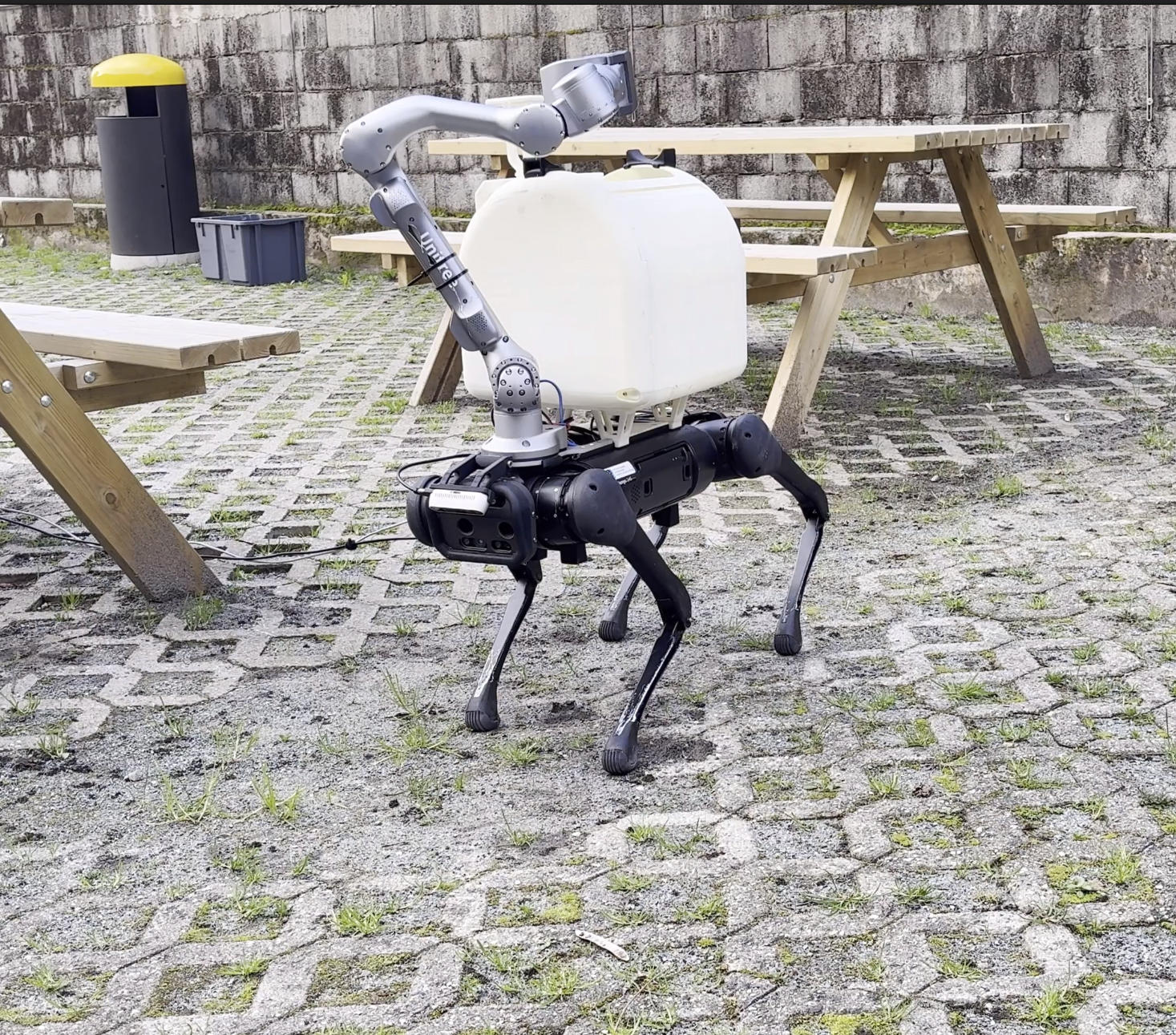} \\[0.3em]

\includegraphics[width=0.328\linewidth]{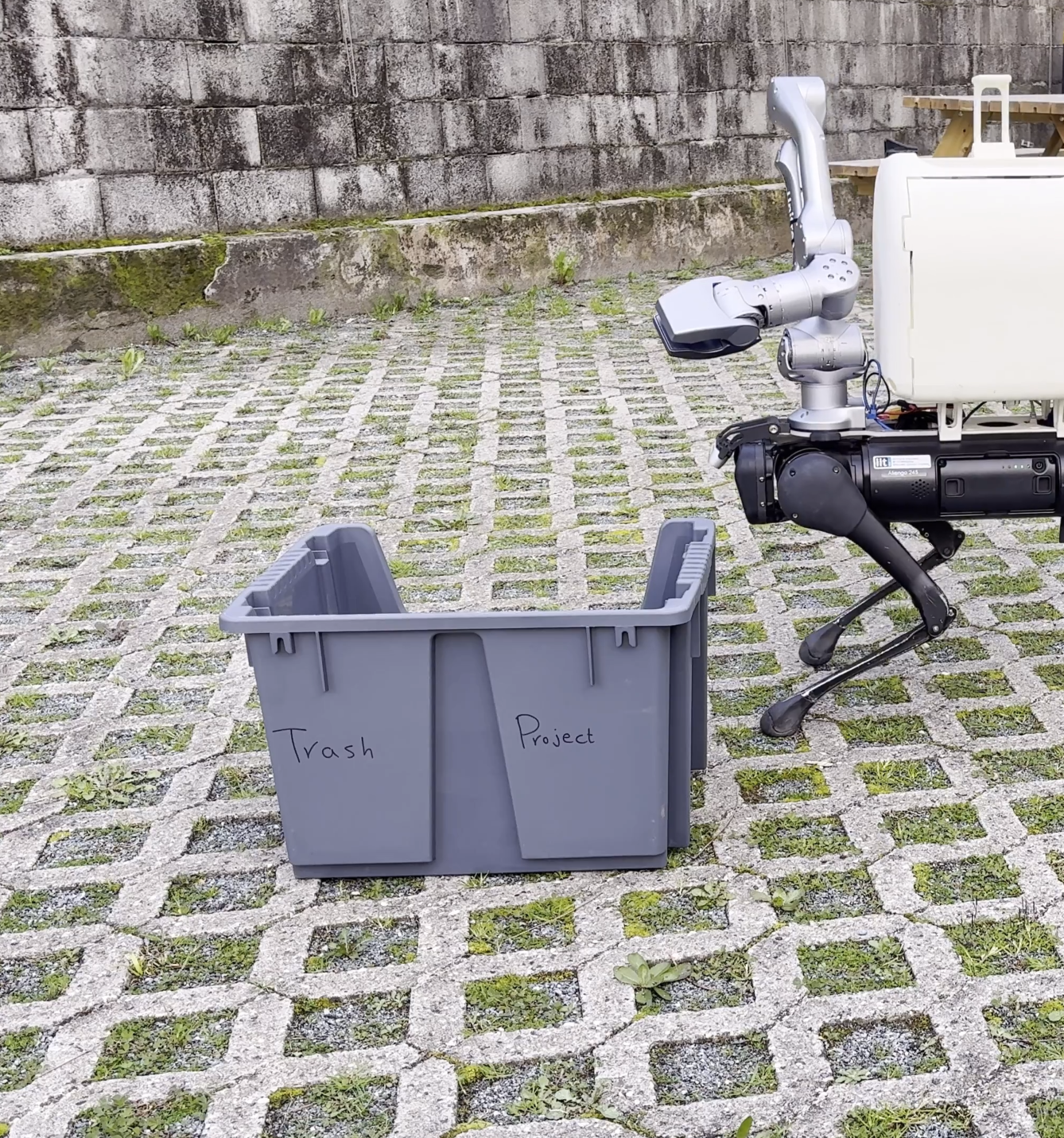}
\includegraphics[width=0.328\linewidth]{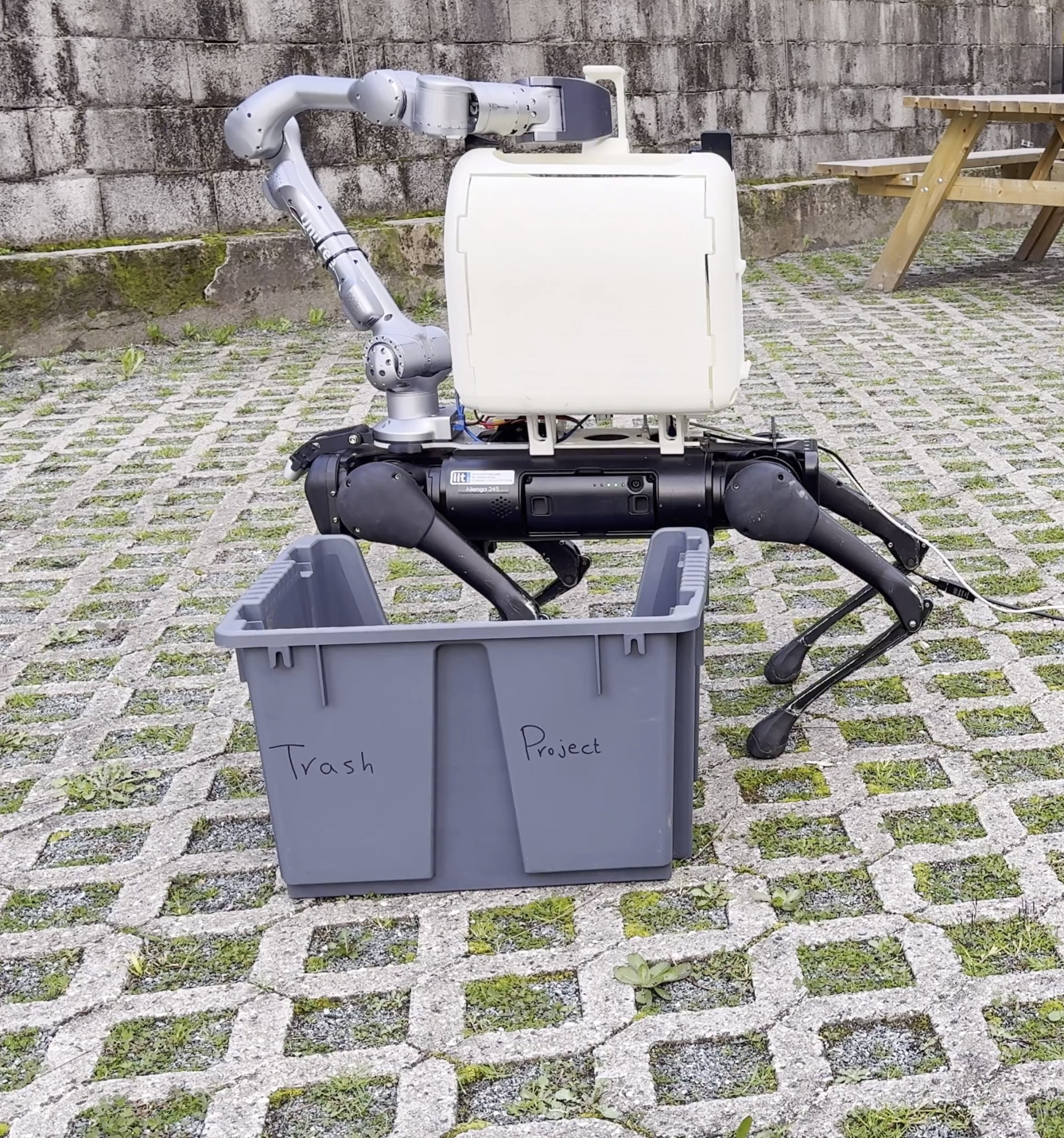}
\includegraphics[width=0.328\linewidth]{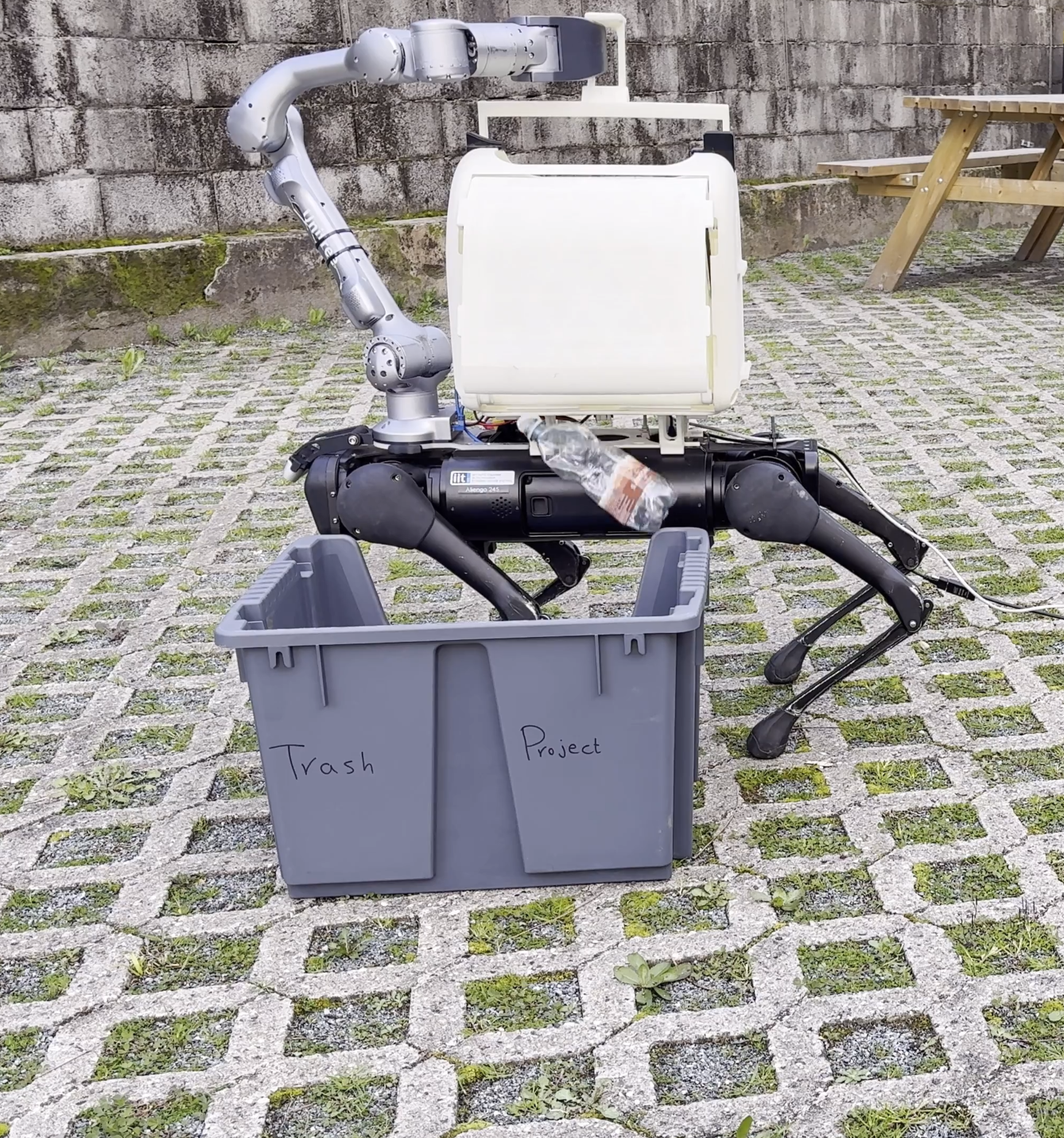}

\caption{
Snapshots of the collection and bin unloading procedures during our experiments. From top-left to bottom-right: the first two rows show a successful grasp and collection of a plastic bottle; the last row shows instead the approach to an external litter box, where the robot can finally unload the collected items at a designated location.
}
\label{fig:experiments}
\end{figure*}

To test our prototype, we conducted litter collection experiments in a typical environment where litter can be commonly found. The selected scenario is a picnic area composed of several benches and flat terrain, where litter may accumulate after meals. Fig. \ref{fig:experiments} shows snapshots from one of the experimental trials.

Starting from the top-left corner of the figure, we illustrate the collection phase of our pipeline. The robot, teleoperated through a joystick, approaches the vicinity of a plastic bottle. A high level layer could substitute the joystick and automate this task. In parallel, the vision module keeps looking for objects and once detected it computes the grasp pose. Once the robot is sufficiently close to the target object, the state machine activates the inverse kinematics module.  At this stage, the robot base is slightly adjusted to improve the grasping configuration, given the relative proximity of the litter item. The IK module then provides the low-level controller with the optimal arm joint configuration required to perform the grasp. After reaching the desired arm configuration, the state machine proceeds by commanding the closure of the gripper to grasp the object.

The collection process then continues: the state machine selects, among the available primitives, the one corresponding to the \textit{box-reaching} action, after which the gripper is opened to release the collected object into the container.

The last three snapshots in Fig. \ref{fig:experiments} illustrate the bin-unloading task. This operation is typically performed after multiple collection cycles. The robot approaches a designated unloading location, where the accumulated litter can later be organized and removed by a human operator. In this phase, the robot is again guided through the joystick, via velocity commands, to reach the unloading area. In this case, the IK module is not required, since the task can be simply completed by concatenating primitives (\textit{box-reaching} $\rightarrow$ \textit{close-gripper} $\rightarrow$ \textit{box-opening}).

The accompanying video shows the complete pipeline in operation across multiple collection attempts.

\subsection{Discussion}
Litter collection can involve multiple challenges that may hinder its success. In this section, we provide a description of these challenges, together with possible solutions and improvements, based on the experience gained while developing this work.

First, litter can sometimes be very difficult to grasp. Consider the plastic bottles analyzed in this manuscript. They can be found smashed, located in difficult-to-reach spots where collisions with the robotic arm are likely, or even partially buried under sand in a beach scenario. These situations challenge different modules developed within this work. Model-based controllers, such as IK, do not reason about contacts. As a consequence, unburying litter may require multiple sequential actions and additional logic, which inevitably tends to fail in complex scenarios. Vision systems also struggle in certain conditions; for example, in the case of smashed bottles, identifying an optimal grasping point becomes significantly more difficult. Finally, objects located in difficult-to-reach areas may require more advanced perception capabilities, which were not developed within the scope of this work.

Several solutions proposed in the literature could address these limitations and are worth investigating for this task. Contact-rich manipulation can be achieved through sampling-based controllers or RL policies \cite{howell2022}. However, these approaches typically require the design and simulation of such scenarios beforehand. Methods for determining optimal grasping points for objects with different topologies have also been proposed in the literature \cite{grasp_segment}. Nevertheless, these methods often require high computational resources, and datasets containing non-ideal or deformed objects are not commonly available online. Finally, difficult-to-reach situations that require advanced perception for collision detection and compliant interaction (for example, using the tip of the gripper near a stair edge to reposition the object before grasping it) could again be addressed with learning-based controllers (VLA) \cite{Black20240AV}. These approaches have demonstrated remarkable performance in this context, but they require extensive training time and computation capabilities. RL, in this case, can again be a viable solution. In the codebase we released with this paper, the reader can find a preliminary implementation of an RL policy for this scenario, which can be integrated to replace the IK module when required.

Even the hardware plays an important role in the task. In this work, the design of the container has maximized the autonomous capability of the system, e.g., allowing the arm to empty it without human intervention. Further modifications to the manipulator, such as a rubber covering for the inner part of the gripper could increase the grasping success, preventing the object from falling while the arm is performing the \textit{box-reaching} action.  

Another important component of a fully autonomous litter collection system is navigation. In this work, navigation was intentionally left outside the scope of the study, by relying instead on joystick commands from a human operator, in order to focus on the design and control of the prototype. Autonomous navigation remains an open problem in the robotics community, and recent approaches, such as \cite{nav}, show promising results that could enable long-term deployment at a city-level scale.

Overall, we believe that locomotion, bin design, and the use of primitives for loading and unloading litter do not currently represent limiting factors for this task and can therefore be adopted as they are in an improved version of the pipeline.

\section{Conclusions}
This work presented BinWalker, a cleaning-oriented robotic platform designed to collect hand-sized litter in unstructured environments using a quadruped base, an onboard manipulator, and an integrated storage container. We demonstrated how the tight integration of mechanical design, hierarchical control, and deep-learning-based computer vision can be a viable solution to perform an autonomous litter collection in a realistic outdoor scenario. The proposed architecture provides a practical blueprint for deploying legged service robots in real-world cleaning applications, highlighting what works and what should be improved in order to make such systems reliable in everyday litter collection tasks.   

\bibliographystyle{IEEEtran}
\bibliography{ref}

\end{document}